%% file: paper.tex
\documentclass[11pt]{uiuc_paper}
\pdfoutput=1
\usepackage{amsmath,amssymb,amsthm,array,flafter}
\input{macro}

\title{Batched Contextual Reinforcement II:\\Does RLVR Need Long Rollout Windows?}
\author[1,*]{Bangji Yang}
\author[2,*]{Hongbo Ma}
\author[1]{Ruihan Guo}
\author[1]{Jiajun Fan}
\author[3]{Xi Zhu}
\author[1,\textdagger]{Ge Liu}
\affiliation[1]{University of Illinois at Urbana-Champaign}
\affiliation[2]{Tsinghua University}
\affiliation[3]{Rutgers University}
\contribution[*]{Equal contribution}
\contribution[\textdagger]{Corresponding author}
\abstract{\input{input/000abstract}}
\date{\today}
\correspondence{Ge Liu, \email{geliu@illinois.edu}}
\hypersetup{
 pdftitle={Batched Contextual Reinforcement II: Does RLVR Need Long Rollout Windows?},
 pdfauthor={Bangji Yang, Hongbo Ma, Ruihan Guo, Jiajun Fan, Xi Zhu, Ge Liu},
 pdfsubject={Efficient reinforcement learning with verifiable rewards}
}

\begin{document}
\maketitle
\input{input/010intro}
\input{input/015windows}
\input{input/030method}
\input{input/035stability}
\input{input/040experiments}
\input{input/020related}
\input{input/050conclusion}

\clearpage
\bibliographystyle{unsrtnat}
\bibliography{main}

\appendix
\clearpage
\input{input/060derivations}
\clearpage
\input{input/070configuration}
\input{input/080details}
\end{document}

%% file: macro.tex
\newtheorem{proposition}{Proposition}
\newtheorem{theorem}{Theorem}
\newcommand{\E}{\mathbb{E}}
\newcommand{\Var}{\operatorname{Var}}
\newcommand{\Cov}{\operatorname{Cov}}
\newcommand{\SNR}{\operatorname{SNR}}

\definecolor{WarmBerry}{HTML}{94344C}
\definecolor{WarmAmber}{HTML}{C38923}
\BeforeBeginEnvironment{proposition}{\begin{samepage}}
\AfterEndEnvironment{proposition}{\end{samepage}}
\BeforeBeginEnvironment{theorem}{\begin{samepage}}
\AfterEndEnvironment{theorem}{\end{samepage}}

%% file: input/000abstract.tex
Long rollout windows make reinforcement learning with verifiable rewards (RLVR) expensive, yet simply shortening them can remove the correctness signal: when every single-problem rollout truncates before an answer, group-relative accuracy advantages vanish. As a technical continuation of Batched Contextual Reinforcement (BCR), we study how concurrent rewards relax this dependence on long windows. Concurrent Constraint Reinforcement (CCR) places three independently verifiable problems in one capped training completion. Correct answers completed before truncation retain credit, while a shared budget encourages shorter useful reasoning without an explicit length penalty. We connect this mechanism to conditional reward- and update-variance analyses and study 4B and 8B mathematical reasoning models. Against single-problem GRPO with a 32k window, CCR with an 8,000-token window reduces tokens per training step from 975,464 to 174,957 on 4B and from 793,832 to 165,788 on 8B, saving 82.1\% and 79.1\%, respectively. An 8B training trace shows increasing accuracy reward alongside shorter completions. Single-problem evaluations provide supporting evidence of retained reasoning quality. These results identify reward-preserving training structure as a practical route to lower per-step cost in short-window RLVR.

%% file: input/010intro.tex
\section{Introduction}

Reinforcement learning with verifiable rewards (RLVR) improves mathematical reasoning by repeatedly generating candidate solutions, checking their answers, and updating the policy~\citep{shao2024deepseekmath,guo2025deepseekr1,cai2025reinforcement}. Long reasoning therefore raises the cost of collecting training experience. The \emph{rollout window} caps each completion. Large windows let models finish long solutions, but permit expensive rollouts throughout training.

Simply shrinking the window can destroy the signal needed to learn. In single-problem GRPO, truncation before an answer yields zero accuracy reward. If every candidate for a prompt is truncated this way, relative accuracy advantages vanish. Repeated across groups, this \emph{reward starvation} can stall learning. Larger windows provide room to finish answers at a higher token budget. This creates the central challenge: \emph{can RLVR retain useful, stable training signals with a smaller rollout budget?}

\begin{figure*}[t]
\centering
\includegraphics[width=\textwidth]{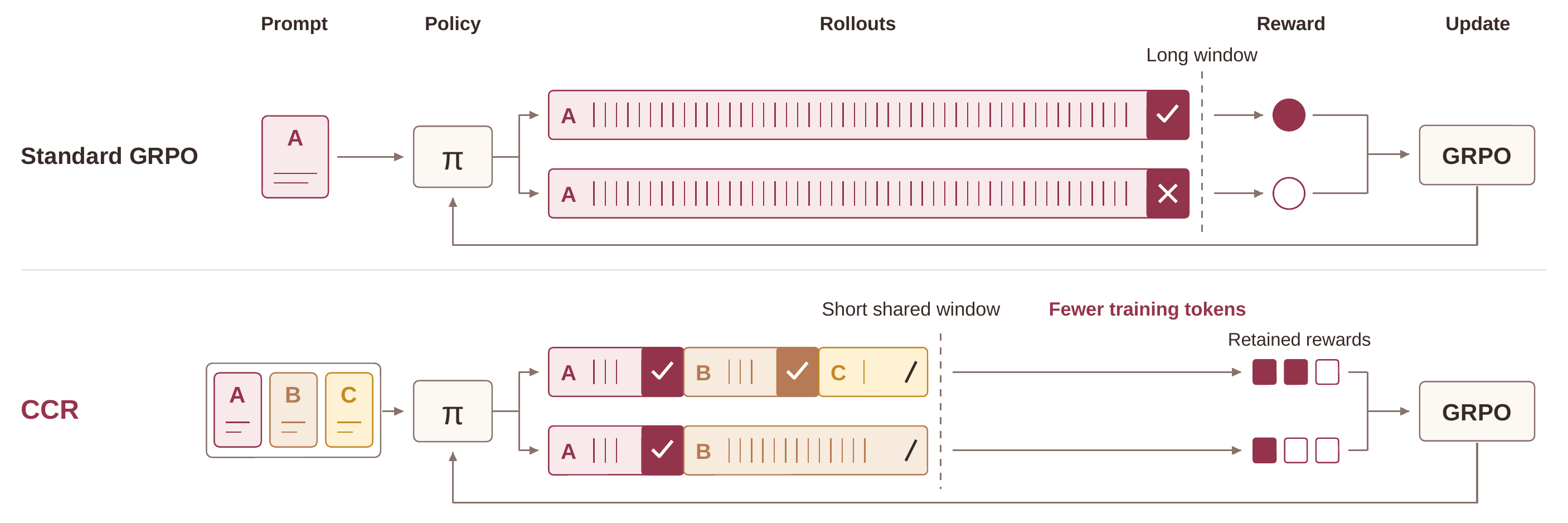}
\caption{\textbf{Standard GRPO versus CCR.} Top: each candidate solves one problem with a long window and receives a terminal correctness reward. Bottom: each CCR candidate attempts grouped problems sequentially under a shorter shared cap. Correct completed answers retain credit despite suffix truncation; correctness is averaged over all assigned problems for the update. Checks, crosses, and slashes denote correct answers, incorrect answers, and truncation. Shorter ribbons illustrate training-token efficiency, not measured ratios. Reward averaging motivates the conditional stability analysis in Section~\ref{sec:stability}.}
\label{fig:algorithm}
\end{figure*}

\emph{Concurrent Constraint Reinforcement} (CCR; Figure~\ref{fig:algorithm}) addresses this challenge by changing the unit of reward. Building on batched contextual reinforcement~\citep{yang2026batched}, it packs independently verifiable problems into one prompt and generates their solutions sequentially under a shared completion cap. Correctness is averaged over all assigned problems. An unfinished suffix contributes zero without erasing earlier correct answers. Candidates can thus receive different rewards even when all are truncated, allowing a short window to retain comparative feedback.

The shared budget connects this feedback to efficiency: unnecessary reasoning on one problem can displace another correct answer. CCR therefore encourages useful compression without an explicit token penalty. Our theory makes the link to stability precise. Different counts of correct completed answers guarantee nonzero centered accuracy advantages despite suffix truncation. Conditional independence yields an $O(1/N)$ reward-noise bound, while an approximate update analysis identifies the non-shared variance reduced by averaging. These guarantees concern the training signal within a feasible regime; they require the stated dependence assumptions and do not guarantee convergence for arbitrary windows.

We evaluate this training design on 4B and 8B models at fixed concurrency $N=3$, using an 8,000-token window against conventional single-problem, 32k-window GRPO. Our evidence combines per-step token cost, an 8B training trace, and ordinary single-problem evaluations. The method retains GRPO and requires no grouped questions at deployment. Cumulative token cost to matched accuracy remains a separate measurement.

Our contributions address three aspects of this objective:
\begin{itemize}
\item \textbf{Reward preservation.} We formalize correctness feedback after suffix truncation and the shared-budget incentive for useful compression.
\item \textbf{Conditional stability.} We establish reward-noise and approximate update-variance bounds, separating signal availability from stability within a feasible window.
\item \textbf{Training efficiency.} CCR reduces tokens per step by 82.1\% on 4B and 79.1\% on 8B, with a learning trace and single-problem evaluations supporting retained reasoning quality.
\end{itemize}

%% file: input/015windows.tex
\section{Rollout Windows and Reward Starvation}
\label{sec:window}

Let $q$ be one mathematical problem and let $y_1,\ldots,y_S$ be $S\geq2$ candidate completions sampled for that prompt. A completion contains at most $B$ generated tokens, including reasoning and answer tokens; prompt tokens are outside this cap. We distinguish this completion limit from the model's total context capacity. Under terminal verification,
\begin{equation}
r_j=\mathbf{1}\{y_j\text{ contains a correct answer}\}.
\end{equation}
Missing or unverifiable answers receive zero. The centered accuracy advantage is
\begin{equation}
a_j=r_j-\frac{1}{S}\sum_{s=1}^{S}r_s.
\label{eq:centered}
\end{equation}
Dividing by a stabilized group standard deviation, as in GRPO, does not change the fact that equal rewards produce zero advantages.

If every candidate reaches the cap before its answer, then $r_j=a_j=0$ for all $j$. This is \emph{accuracy-reward starvation}: those candidates cannot be ranked by correctness. Groups with different rewards retain task-directed feedback; format and KL terms can contribute separately to the update. The issue addressed here is the missing correctness signal in groups whose answers are all truncated.

A longer window can let more trajectories reach verifiable answers. The maximum window is an upper bound, however, rather than the number of tokens actually consumed. Early stopping, problem difficulty, and the policy all affect realized rollout length. We therefore measure tokens per training step for each configuration. This describes the realized token workload under each training configuration. We use a 32k completion window for the standard GRPO comparison.

%% file: input/030method.tex
\section{Concurrent Rewards for Short Windows}
\label{sec:method}

\subsection{Training unit and reward}

For a training group $G=(q_1,\ldots,q_N)$, the policy generates one ordered completion containing the solutions to all $N$ problems, subject to $L(y)\leq B$. We use $N=3$ in the reported CCR experiments. This is concurrency within a completion, not GPU minibatching: all three problems compete for the same generated-token budget.

For each $G$, sample $S$ joint completions. Let $X_{i,j}$ indicate that completion $j$ contains a complete, correct answer to problem $i$. Missing, malformed, or incorrect answers have $X_{i,j}=0$. Define
\begin{align}
r_{\mathrm{acc},j}&=\frac{1}{N}\sum_{i=1}^{N}X_{i,j},\label{eq:acc}\\
R_j&=w_{\mathrm{acc}}r_{\mathrm{acc},j}
       +w_{\mathrm{fmt}}r_{\mathrm{fmt},j},\label{eq:reward}\\
\widetilde A_j&=R_j-\frac{1}{S}\sum_{s=1}^{S}R_s.\label{eq:total-adv}
\end{align}
The reported CCR reward weights are $w_{\mathrm{acc}}=2$ and $w_{\mathrm{fmt}}=1$. The format term checks the required answer structure. Crucially, accuracy is computed independently for each recoverable answer: a malformed or truncated suffix must not invalidate a correctly completed prefix. The denominator remains $N$, including unfinished problems, so omitting difficult answers cannot inflate the average over attempted problems.

Group-relative comparisons are made across candidates for the \emph{same} $G$. In particular, a candidate solving one answer is not compared against a candidate generated for an unrelated set of easier questions. CCR supplies these rewards to a GRPO-style policy update~\citep{shao2024deepseekmath}. For clarity, the idealized, unnormalized policy-gradient term is
\begin{equation}
\widehat g=\frac{1}{S}\sum_{j=1}^{S}\widetilde A_j
\nabla_\theta\log\pi_\theta(y_j\mid G).
\label{eq:gradient}
\end{equation}
The training implementation may include probability-ratio clipping, advantage scaling, and KL regularization. Equation~\eqref{eq:gradient} isolates the accuracy-driven signal for the analysis below.

\paragraph{Operational training procedure.}
CCR samples joint candidates, verifies each recoverable answer, and computes within-group advantages from the combined accuracy and format reward. Verification stops at the last recoverable answer, while normalization includes all assigned problems. Appendix~\ref{app:procedure} gives the procedure and an illustrative answer-recovery interface.

\subsection{Completed answers survive truncation}
\label{sec:partial}

\begin{proposition}[Accuracy signal after suffix truncation]
\label{prop:signal}
Let $C_j=\sum_iX_{i,j}$ be the number of correct completed answers in candidate $j$. The centered accuracy advantages satisfy
\begin{equation}
a_j=\frac{C_j-\bar C}{N},\qquad
\bar C=\frac{1}{S}\sum_{s=1}^{S}C_s.
\label{eq:prefix-adv}
\end{equation}
For $N=1$, truncation before every candidate's sole answer gives $a_j=0$ for all $j$. For $N>1$, any variation in $C_j$ gives a nonzero accuracy advantage, even if every candidate has an unfinished suffix.
\end{proposition}

Substituting $r_{\mathrm{acc},j}=C_j/N$ into Eq.~\eqref{eq:centered} proves the result. In Table~\ref{tab:toy}, four candidates complete zero, one, two, and one correct answers. Although none finishes the three-problem group, their rewards yield both positive and negative advantages.

\begin{table}[htbp]
\centering
\small
\begin{tabular}{lccc}
\toprule
Candidate & Correct prefix & $r_{\mathrm{acc}}$ & $a_j$\\
\midrule
$y_1$ & 0 answers & $0$ & $-1/3$\\
$y_2$ & 1 answer & $1/3$ & $0$\\
$y_3$ & 2 answers & $2/3$ & $+1/3$\\
$y_4$ & 1 answer & $1/3$ & $0$\\
\bottomrule
\end{tabular}
\caption{Illustrative $N=3$, $S=4$ group. All candidates truncate before completing the group. Their mean accuracy reward is $1/3$; completed prefix answers preserve relative feedback.}
\label{tab:toy}
\end{table}

Preserving positive rewards alone is insufficient: candidates must differ in reward to create a group-relative signal. CCR can still starve if the window is too short to finish the first answer, or if every candidate completes the same number of correct answers. Moreover, retaining a parsed reward is useful only if the trainer also retains the corresponding policy-loss contribution. Discarding every truncated completion would remove the very feedback the construction is intended to preserve. The mechanism therefore requires prefix-tolerant verification and a compatible truncation mask.

\paragraph{How often is a group informative?}
The reward resolution also admits an exact finite-sample description. For a fixed problem group and cap, let $p_c=\Pr(C_j=c\mid G,B)$ for $c=0,\ldots,N$. If the $S$ candidates are sampled independently, then
\begin{equation}
\Pr(\exists j:a_j\ne0\mid G,B)
=1-\sum_{c=0}^{N}p_c^S.
\label{eq:informative}
\end{equation}
Indeed, all centered accuracy advantages vanish exactly when all candidates have the same correct-answer count. The all-zero case has probability $p_0^S$, but equal positive rewards can also be uninformative. This expression requires independence across sampled candidates, not across problems inside a candidate.

Concurrency supplies additional possible reward levels, but additional levels help only if the policy places probability on them. When $p_0=1$, Eq.~\eqref{eq:informative} is zero regardless of $N$ or $S$. Conversely, spreading mass across distinct completed-answer counts can preserve useful comparisons even when every rollout is truncated. This identifies a concrete diagnostic for short-window training: log the fraction of groups with nonzero accuracy-reward variance, rather than using average reward or truncation rate alone as a proxy for the learning signal.

\subsection{Why the same constraint encourages compression}

Let $K(y)$ be the completed prefix length in an ordered trajectory. For a finished problem, $\ell_i(y)$ counts its reasoning and answer tokens; set $\ell_i(y)=\infty$ when problem $i$ is unfinished. Then
\begin{equation}
K(y)=\max\left\{k\leq N:\sum_{i=1}^{k}\ell_i(y)\leq B\right\},
\end{equation}
where the maximum is zero if the first answer is unfinished. Unfinished problems contribute zero to Eq.~\eqref{eq:acc}. Under this simplified prefix model, reward is the sum of correctness indicators for completed problems divided by $N$.

\begin{proposition}[Opportunity cost of reasoning tokens]
\label{prop:compression}
Suppose concise and verbose trajectories complete $K_c>K_v$ problems and have equal success probabilities on their shared completed prefix. If $p_i$ is the success probability on each additional completed problem, then
\begin{equation}
\E[r_c-r_v]=\frac{1}{N}\sum_{i=K_v+1}^{K_c}p_i\geq0.
\label{eq:compression}
\end{equation}
The inequality is strict when an additional problem has positive success probability.
\end{proposition}

The statement does not reward brevity by itself: it assumes that shortening preserves correctness on the shared prefix. Useful extra reasoning remains valuable. If an edit increases problem $i$'s success probability by $\Delta p_i$ without displacing an answer, its expected accuracy reward gain is $\Delta p_i/N$. If it displaces a set $D$ of otherwise completed answers, the gain becomes
\begin{equation}
\Delta\E[r_{\mathrm{acc}}]
=\frac{1}{N}\left(\Delta p_i-\sum_{k\in D}p_k\right).
\label{eq:opportunity}
\end{equation}
In contrast, an explicit reward $r_{\mathrm{acc}}-\lambda L$ subtracts $\lambda\Delta\ell_i$ for every added token. CCR imposes no such fixed tax. It optimizes expected correctness under a hard shared budget, making unnecessary reasoning costly when that reasoning crowds out another answer.

These effects support a common training mechanism. Completed prefixes offer a route to comparative feedback before all problems fit within the cap. When that feedback favors trajectories that finish more correct answers, the policy has an incentive to allocate its limited tokens more effectively. Additional useful reasoning can increase both reward and length; the shared budget favors productive allocation rather than uniformly shorter responses.

%% file: input/035stability.tex
\section{Stability Within the Feasible Regime}
\label{sec:stability}

Partial credit explains how a short-window group can remain informative. We next analyze the complementary variance-reduction mechanism: when multiple answers are feasible, averaging their correctness can reduce observation noise, and averaging problem-level updates can reduce non-shared variance. We state the relevant independence and coupling assumptions explicitly.

\subsection{Reward noise and accuracy advantages}

For completion $j$, let $p_{i,j}$ be the latent probability that problem $i$ yields an observed correct answer and set $\bar p_j=N^{-1}\sum_i p_{i,j}$. Write
\begin{equation}
r_{\mathrm{acc},j}=\bar p_j+\epsilon_j.
\end{equation}
We analyze the \emph{accuracy component} of the centered advantage.

\begin{theorem}[Conditional reward-noise reduction]
\label{thm:reward}
If $X_{1,j},\ldots,X_{N,j}$ are independent Bernoulli variables conditional on their latent probabilities, then
\begin{equation}
\Var(\epsilon_j\mid\{p_{i,j}\})
=\frac{1}{N^2}\sum_i p_{i,j}(1-p_{i,j})
\leq\frac{1}{4N}.
\label{eq:noise}
\end{equation}
With independent candidate completions, the centered accuracy advantage decomposes as
\begin{equation}
a_j=(\bar p_j-\overline{\bar p})
 +(\epsilon_j-\bar\epsilon),
\label{eq:signal-noise}
\end{equation}
and its noise term has variance $O(1/N)$ for fixed $S$.
\end{theorem}

A proof and the exact baseline-centering factor appear in Appendix~\ref{app:reward}. A linear signal-to-noise improvement additionally requires nonvanishing latent-quality variation across candidates and reward-noise variance of order $1/N$. Under these stronger conditions, $\SNR(a)=\Theta(N)$.

The independence condition also matters. For correlated correctness indicators, the conditional variance contains an additional term,
\begin{equation}
\frac{2}{N^2}\sum_{i<k}\Cov(X_{i,j},X_{k,j}\mid p).
\label{eq:covariance}
\end{equation}
A shared token cap and sequential generation can induce exactly this dependence. The covariance term characterizes how dependence modifies the stability benefit of reward averaging.

\subsection{Averaged updates and residual coupling}

An analogous calculation applies if the accuracy-driven joint update is well approximated by an average of problem-level updates. Let $g_i$ denote the centered update associated with problem $i$ and suppose
\begin{equation}
g_{\mathrm{acc}}\approx\frac{w_{\mathrm{acc}}}{N}\sum_i g_i.
\label{eq:avg-update}
\end{equation}
For a unit direction $u$, write $Z_i=u^\top g_i$. If all $Z_i$ have variance $\sigma_u^2$ and pairwise correlation $\rho_u$, then
\begin{equation}
\Var(u^\top g_{\mathrm{acc}})
\approx w_{\mathrm{acc}}^2\sigma_u^2
\left(\rho_u+\frac{1-\rho_u}{N}\right).
\label{eq:update-var}
\end{equation}
At fixed covariance structure, averaging reduces the non-shared component when $\rho_u<1$; the shared component remains. Appendix~\ref{app:update} derives the cross-problem residual omitted by Eq.~\eqref{eq:avg-update}. A shared budget does not by itself make that residual small.

\subsection{Choosing a feasible shared budget}

Stability and reward preservation both depend on the availability of usable answers. If a model needs roughly $b$ tokens per solution, then fitting all $N$ solutions requires a budget on the order of $Nb$, although actual lengths vary by problem and trajectory. Increasing $N$ can therefore make the cap infeasible and destroy the signal it was intended to enrich. We fix $N=3$ and study its training behavior for two backbones.

%% file: input/040experiments.tex
\section{Experiments}
\label{sec:experiments}

\subsection{Training and evaluation setup}

We study Qwen3-4B-Thinking and Qwen3-8B-ODA-Math with training concurrency $N=3$ and a shared completion window of 8,000 tokens. The experiments use a reported subset of 3,000 problems from DeepMath-103K~\citep{he2026deepmath}, a TRL-based training stack~\citep{vonwerra2020trl}, and eight NVIDIA A100 GPUs. CCR combines accuracy and format rewards as in Eq.~\eqref{eq:reward}, with no explicit length penalty or auxiliary process reward. Appendix~\ref{app:reporting} summarizes the configuration and log-processing procedure.

For training cost, we compare CCR with standard single-problem GRPO using a 32k completion window, matching the remaining training parameters and hardware. Fewer than 10\% of the 8B baseline completions are truncated, so the long-window baseline operates with a low truncation rate. We measure tokens consumed per training step under each configuration.

For training dynamics, we use the 8B accuracy-reward and completion-length logs over steps 1--1,000. For downstream quality, we evaluate the base and CCR checkpoints on nine mathematical reasoning benchmarks. Every evaluation prompt contains one problem ($N_{\mathrm{eval}}=1$): concurrent training does not require grouped inference at deployment. The base checkpoints in these evaluations are the starting policies, distinct from the trained GRPO cost baseline.

\subsection{Training-token efficiency at 4B and 8B}

\begin{figure}[htbp]
\centering
\includegraphics[width=\textwidth]{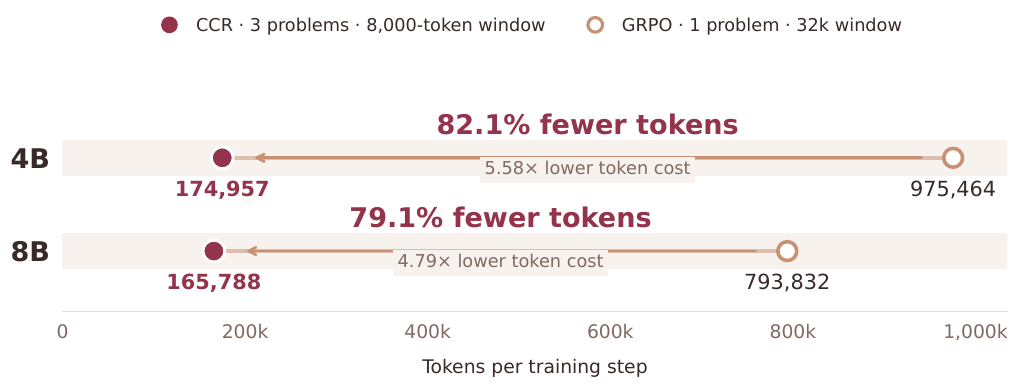}
\caption{\textbf{CCR reduces tokens per training step by 82.1\% on 4B and 79.1\% on 8B.} Filled berry markers denote CCR with three problems and an 8,000-token shared window; hollow clay markers denote single-problem GRPO with a 32k window. The common linear axis shows the absolute per-step counters. Both comparisons use eight A100 GPUs and matched remaining training parameters.}
\label{fig:cost}
\end{figure}

Figure~\ref{fig:cost} compares the measured per-step token workload. On 4B, CCR reduces token consumption from 975,464 to 174,957, an 82.1\% reduction. On 8B, it reduces consumption from 793,832 to 165,788, a 79.1\% reduction. The baseline-to-CCR ratios are 5.58 and 4.79, respectively: CCR uses approximately one fifth of the baseline's tokens per step on both backbones.

This comparison captures the combined training design: multiple verifiable answers share a smaller completion budget. The window is an upper bound on generation, while the per-step counter records realized token use. Prefix credit makes the restricted budget useful for learning, and the shared cap creates pressure to allocate those tokens to answers that can earn reward.

\subsection{Reward improves while completions shorten}

\begin{figure}[htbp]
\centering
\includegraphics[width=\textwidth]{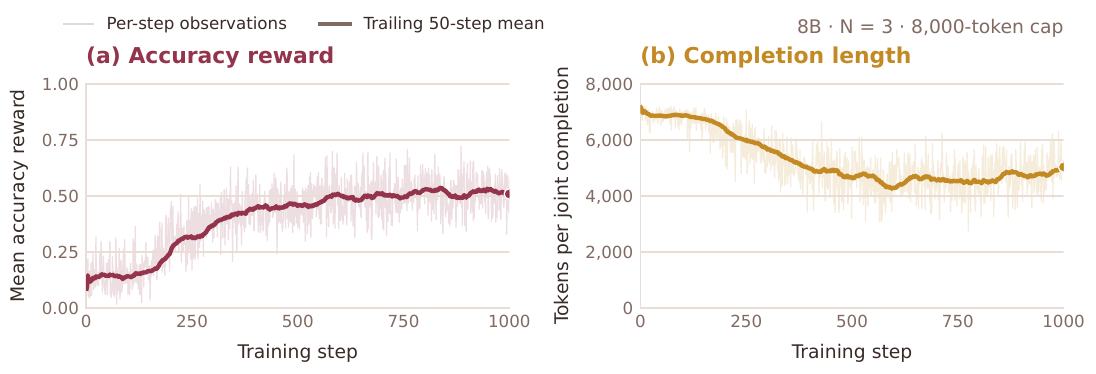}
\caption{\textbf{Accuracy improves as training completions shorten.} The 8B CCR run uses $N=3$ and an 8,000-token shared cap. Faint traces show raw observations over steps 1--1,000; solid traces show trailing 50-step means of accuracy reward and joint-completion length.}
\label{fig:dynamics}
\end{figure}

Figure~\ref{fig:dynamics} shows the 8B CCR trajectory. Faint lines display the raw per-step observations, and solid lines show a trailing 50-step mean. Accuracy reward rises from 0.142 over the first 50 steps to 0.509 over the last 50 steps. Over the same intervals, mean joint-completion length falls from 6,867 to 5,034 tokens, a 26.7\% decrease.

The completion cap stays fixed throughout this run, so the shortening reflects changes in the policy's generated trajectories rather than a progressively smaller configured window. Completion length declines through much of training and increases modestly near the end, while accuracy reward continues to improve. This pattern is consistent with the opportunity-cost mechanism in Eq.~\eqref{eq:opportunity}: extra reasoning remains valuable when it helps produce correct answers within the shared budget.

The cost comparison and learning trajectory describe complementary aspects of efficiency. The former measures the token workload of the short-window configuration; the latter shows continued learning as the policy uses that budget more compactly. Proposition~\ref{prop:signal} explains how completed prefixes can preserve the relative feedback needed for this process.

\subsection{Reasoning quality after short-window training}

Table~\ref{tab:full-benchmarks} reports single-problem evaluation results. Average accuracy increases from 75.3\% to 77.7\% on 4B and from 68.8\% to 69.7\% on 8B, with variation across individual benchmarks. These results show that the CCR checkpoints retain useful mathematical reasoning after short-window training. Evaluation lengths also decrease, providing supporting context for the training-efficiency results.

\begin{table}[htbp]
\centering
\small
\renewcommand{\arraystretch}{1.05}
\setlength{\tabcolsep}{5pt}
\begin{tabular}{lrrrrrrrr}
\toprule
& \multicolumn{4}{c}{Qwen3-4B-Thinking} & \multicolumn{4}{c}{Qwen3-8B-ODA-Math}\\
\cmidrule(lr){2-5}\cmidrule(lr){6-9}
& \multicolumn{2}{c}{Accuracy (\%)} & \multicolumn{2}{c}{Eval. tokens} & \multicolumn{2}{c}{Accuracy (\%)} & \multicolumn{2}{c}{Eval. tokens}\\
Benchmark & Base & CCR & Base & CCR & Base & CCR & Base & CCR\\
\midrule
AIME24 & 83.3 & 83.3 & 18,082 & 14,539 & 63.3 & 70.0 & 16,549 & 13,680\\
AIME25 & 70.0 & 83.3 & 20,773 & 17,498 & 66.7 & 60.0 & 17,832 & 13,693\\
AMC23 & 97.5 & 100.0 & 10,457 & 7,128 & 95.0 & 97.5 & 6,988 & 4,516\\
MATH-500 & 97.0 & 97.4 & 5,136 & 3,173 & 95.4 & 94.4 & 4,010 & 2,773\\
Minerva & 66.2 & 66.5 & 4,576 & 3,656 & 61.0 & 62.9 & 4,910 & 3,674\\
Olympiad-Bench & 83.3 & 85.2 & 13,069 & 10,717 & 75.0 & 77.3 & 10,767 & 7,654\\
BRUMO25 & 66.7 & 73.7 & 19,133 & 13,782 & 73.3 & 73.3 & 15,399 & 11,175\\
CMIMC25 & 60.0 & 60.0 & 23,286 & 18,173 & 42.5 & 45.0 & 21,604 & 14,993\\
HMMT25 & 53.3 & 50.0 & 23,515 & 18,564 & 46.7 & 46.7 & 19,323 & 16,862\\
\midrule
Reported average & 75.3 & 77.7 & 15,336 & 11,914 & 68.8 & 69.7 & 13,042 & 9,891\\
\bottomrule
\end{tabular}
\caption{\textbf{Reasoning quality after short-window training.} CCR is trained with $N=3$ and evaluated on single-problem prompts. Accuracy is final-answer correctness (\%); tokens are evaluation lengths. Base denotes the starting checkpoint.}
\label{tab:full-benchmarks}
\end{table}

%% file: input/020related.tex
\section{Related Work}

\paragraph{RLVR and rollout design.}
GRPO uses relative rewards among candidate completions to train mathematical reasoning policies~\citep{shao2024deepseekmath}; DeepSeek-R1 demonstrates the broader role of verifiable RL in eliciting reasoning~\citep{guo2025deepseekr1}. DAPO addresses long-chain RL training with dynamic sampling and overlong reward shaping, among other changes~\citep{yu2025dapo}. Verifier noise can corrupt these rewards even when answers are checkable~\citep{cai2025reinforcement}; SocraticPO adds corrective teacher guidance to failed rollouts~\citep{liu2026socraticpopolicyoptimizationinteractive}. Heterogeneous agent collaboration offers another efficiency route by sharing verified rollouts across agents~\citep{zhang2026heterogeneous}. These approaches highlight that sampling and reward handling affect the usefulness of generated trajectories. CCR changes the number of independently verifiable answers available within a trajectory, so an unfinished suffix need not eliminate all accuracy feedback.

\paragraph{Grouped reasoning and implicit compression.}
Batched contextual reinforcement already introduced multiple problems in a shared context as a structural incentive for concise reasoning~\citep{yang2026batched}. We build on that construction, rather than claiming a new optimizer or the first use of grouped prompts. Our focus is its consequence for RLVR training: the coupling between completion caps, reward availability, and measured step cost. The conditional stability analysis supports this motivation without turning fixed-concurrency results into a task-scaling claim.

\paragraph{Reasoning and inference compute.}
Chain-of-thought prompting and self-consistency improve reasoning by eliciting or sampling intermediate trajectories~\citep{wei2022chainofthought,wang2023selfconsistency}. Test-time compute allocation studies how additional inference can improve individual predictions~\citep{snell2025scaling}. Related efficiency work studies when reasoning should stop~\citep{huang2026does} and how latent guidance can transfer reasoning to smaller models~\citep{chen2026latent}. In medical vision-language tasks, adaptive causal reflection and latent diagnostic memory provide distinct ways to shape reasoning trajectories~\citep{lin2026models,zhu2026medsynapsevbridgingvisualperception}. Here the scarce resource is repeated on-policy generation during training. Shorter evaluation traces are useful supporting observations, but our primary measurements concern training steps under a restricted rollout window.

\paragraph{Other feedback and evaluation settings.}
Robust preference optimization handles noisy preference labels~\citep{liang2025ropo}; positive-unlabeled RL distillation constructs preference signals for small locally deployed models~\citep{kou2026positiveunlabeled}; and am-ELO addresses ranking stability in arena-based evaluation~\citep{liu2025elo}. These studies concern different feedback or evaluation regimes from CCR's verified multi-problem rewards.

%% file: input/050conclusion.tex
\section{Conclusion}

BCR II develops the training-efficiency perspective of batched contextual reinforcement. CCR preserves credit for completed answers under suffix truncation and rewards productive use of a shared token budget. Our analysis connects prefix credit to group-relative feedback and reward averaging to conditional stability. With $N=3$ and an 8,000-token window, CCR uses 82.1\% fewer tokens per step on 4B and 79.1\% fewer on 8B than single-problem, 32k-window GRPO. Increasing reward, shorter training completions, and retained single-problem reasoning quality support a common conclusion: structuring rollouts around multiple verifiable answers can make short-window RLVR effective and efficient.

%% file: input/060derivations.tex
\section{Derivations and Assumptions}
\label{app:derivations}

\subsection{Prefix rewards and opportunity cost}

For Proposition~\ref{prop:signal}, substituting $r_j=C_j/N$ into the centered advantage gives $a_j=(C_j-\bar C)/N$. If some $C_j$ differ, they cannot all equal their mean, so some advantages are nonzero. In fact, the finite-group accuracy variance is
\begin{equation}
\frac{1}{S}\sum_j a_j^2
=\frac{1}{N^2 S}\sum_j(C_j-\bar C)^2.
\end{equation}
This equality makes the necessary reward-diversity condition explicit. It concerns the accuracy component; the total reward also contains the format term.

For Proposition~\ref{prop:compression}, the expected correctness contributions on the common completed prefix cancel. Only the additional completed problems remain, giving Eq.~\eqref{eq:compression}. For Eq.~\eqref{eq:opportunity}, add the expected gain $\Delta p_i/N$ and subtract the expected correctness of each displaced answer divided by $N$. These are comparisons of expected rewards under the stated preservation assumptions.

\subsection{Reward variance and centering}
\label{app:reward}

Conditional independence gives
\begin{align}
\E[r_j\mid p]&=\bar p_j,\\
\Var(r_j\mid p)&=\frac{1}{N^2}\sum_i p_{i,j}(1-p_{i,j}).
\end{align}
Each Bernoulli variance is at most $1/4$, proving Eq.~\eqref{eq:noise}. With
\begin{equation}
\nu_j=\frac{1}{N}\sum_i p_{i,j}(1-p_{i,j}),
\end{equation}
the conditional noise variance is $\nu_j/N$. Under independent candidate noise, a baseline computed from the same $S$ samples yields
\begin{align}
&\Var(\epsilon_j-\bar\epsilon\mid p)\nonumber\\
&\quad=\frac{1}{N}\left[
\left(1-\frac{2}{S}\right)\nu_j
+\frac{1}{S^2}\sum_s\nu_s\right].
\label{eq:centered-noise}
\end{align}
It is $O(1/N)$ for fixed $S\geq2$. When candidate noise has the same unconditional variance $\nu/N$, this reduces to $(1-1/S)\nu/N$.

If candidates' latent mean qualities are independent and identically distributed with variance $\sigma_p^2$, their centered signal variance is $(1-1/S)\sigma_p^2$. The conditional noise has zero mean, so it is uncorrelated with the latent signal. In this setting,
\begin{equation}
\SNR(a)=\frac{N\sigma_p^2}{\nu}.
\end{equation}
Thus linear order $\Theta(N)$ requires both $\sigma_p^2=\Theta(1)$ and $\nu=\Theta(1)$, with positive lower bounds. A noise upper bound alone only gives a one-sided statement when the signal remains nonzero. If all rewards coincide because all answers truncate, there is no informative latent correctness signal to which this argument can apply. Conditional covariance between problem outcomes adds Eq.~\eqref{eq:covariance} and can prevent $1/N$ decay.

\subsection{Update averaging and cross-problem terms}
\label{app:update}

Write the score of candidate $j$ as
\begin{equation}
s_j=\nabla_\theta\log\pi_\theta(y_j\mid G)
=\sum_{k=1}^{N}s_{k,j}+e_j.
\end{equation}
Here $s_{k,j}$ denotes the component associated with problem $k$ and $e_j$ collects any residual under the chosen segmentation. Define centered correctness indicators
\begin{equation}
a_{i,j}=X_{i,j}-\frac{1}{S}\sum_s X_{i,s},
\end{equation}
so the centered accuracy advantage is $N^{-1}\sum_i a_{i,j}$. Its unnormalized gradient contribution becomes
\begin{align}
g_{\mathrm{acc}}
&=\frac{w_{\mathrm{acc}}}{NS}\sum_{i,j}a_{i,j}s_j\nonumber\\
&=\frac{w_{\mathrm{acc}}}{N}\sum_i g_i+C_{\mathrm{cross}}+C_{\mathrm{res}},
\end{align}
where
\begin{align}
g_i&=\frac{1}{S}\sum_j a_{i,j}s_{i,j},\\
C_{\mathrm{cross}}&=\frac{w_{\mathrm{acc}}}{NS}\sum_j\sum_{i\ne k}a_{i,j}s_{k,j},\\
C_{\mathrm{res}}&=\frac{w_{\mathrm{acc}}}{NS}\sum_{i,j}a_{i,j}e_j.
\end{align}
Equation~\eqref{eq:avg-update} assumes both residual contributions are small. Sequential conditioning and a shared budget may violate this assumption; the constraint alone does not suppress cross terms.

For the averaged component, if $Z_i=u^\top g_i$ has common variance $\sigma_u^2$ and common pairwise correlation $\rho_u$, then
\begin{align}
\Var\!\left(\frac{1}{N}\sum_i Z_i\right)
&=\frac{N\sigma_u^2+N(N-1)\rho_u\sigma_u^2}{N^2}\nonumber\\
&=\sigma_u^2\left(\rho_u+\frac{1-\rho_u}{N}\right).
\end{align}
Multiplying by $w_{\mathrm{acc}}^2$ yields Eq.~\eqref{eq:update-var}. The equal-correlation model must have a valid covariance matrix. This comparison holds $\rho_u$ fixed and isolates the averaged accuracy-gradient component. Format-gradient effects, normalization, and clipping enter through the full GRPO update.

%% file: input/070configuration.tex
\section{Experimental Configuration and Log Processing}
\label{app:reporting}

Table~\ref{tab:config} summarizes the settings used for the comparisons. The training-token measurements use the recorded per-step counters for each method with matched remaining training parameters and hardware.

\begin{table}[ht]
\centering
\small
\begin{tabular}{@{}>{\raggedright\arraybackslash}p{0.40\columnwidth}>{\raggedright\arraybackslash}p{0.54\columnwidth}@{}}
\toprule
Item & Setting\\
\midrule
CCR backbones & 4B-Thinking; 8B-ODA-Math\\
Problems/completion & $N=3$\\
CCR completion cap & 8,000 tokens (both models)\\
GRPO cost baselines & 4B and 8B, one problem, 32k cap\\
Hardware & 8 NVIDIA A100 GPUs\\
CCR reward weights & Accuracy 2; format 1\\
Training source & DeepMath-103K\\
Training subset (reported) & 3,000 problems\\
Evaluation concurrency & One problem per prompt\\
Training implementation & TRL-based stack\\
\bottomrule
\end{tabular}
\caption{Training and evaluation configuration.}
\label{tab:config}
\end{table}

\paragraph{Training dynamics.}
The two W\&B exports each contain 1,008 observations with matching global-step identifiers. We join the accuracy-reward and mean-completion-length columns by global step and select steps 1--1,000. For observations $z_t$, the displayed trailing mean is
\begin{equation}
\bar z_t=\frac{1}{\min(t,50)}\sum_{u=\max(1,t-49)}^{t}z_u.
\end{equation}
The smoother uses the available prefix for the first 49 steps. First-50 and last-50 summaries are arithmetic means of the raw observations. The faint traces in Figure~\ref{fig:dynamics} display the raw values from this run. Source CSVs, plotting scripts, and data hashes accompany the figure assets.

%% file: input/080details.tex
\clearpage
\section{Training Procedure and Reward Interface}
\label{app:procedure}

The following procedure specifies the CCR training structure. Candidate count, optimizer settings, and decoding hyperparameters are inputs to the training loop.

\begin{enumerate}
\item Form an ordered group $G=(q_1,\ldots,q_N)$ of independently verifiable problems; use $N=3$ for the reported runs.
\item Build one prompt with distinct problem identifiers and request an ordered sequence of solutions and recoverable answers.
\item Sample $S$ candidate completions for this same prompt, each with the shared generation cap $B=8{,}000$.
\item For each candidate, extract every completed answer that can be unambiguously assigned to its problem. Set the correctness indicator to zero for missing or incorrect answers. An unfinished suffix does not revoke earlier valid answers.
\item Compute $r_{\mathrm{acc},j}=N^{-1}\sum_i X_{i,j}$, retain the fixed denominator $N$, and add the format term with the reported reward weights.
\item Compute group-relative advantages among the candidates for $G$ and apply the configured GRPO loss. Loss masking must retain the policy contribution associated with the reward-bearing prefix of a truncated trajectory.
\item Repeat with the updated policy. Record completion length, per-step token consumption, and accuracy-reward statistics separately.
\end{enumerate}

\paragraph{Illustrative prompt interface.}
A minimal interface presents \texttt{Problem 1}, \texttt{Problem 2}, and \texttt{Problem 3} and asks for a separately identifiable answer after each solution. For example, \texttt{Answer 1:} followed by a boxed expression identifies the first answer. This illustrates the answer-recovery interface. The implementation should specify delimiters, verifier normalization, ambiguous-answer handling, and how the final answer is associated with each problem.

\paragraph{Partial credit and format checking.}
The format reward and the accuracy reward serve distinct roles. A suffix that violates a formatting convention may lose format credit while leaving earlier correctness indicators valid. Requiring the entire response to pass a format gate before evaluating any answer would reintroduce all-or-nothing terminal feedback. Similarly, dividing by the number of attempted answers would allow the policy to inflate reward by omitting difficult problems; the fixed denominator avoids that incentive.

\paragraph{Truncation and the policy loss.}
A verifier that preserves prefix credit is insufficient if a trainer subsequently masks the whole truncated response. Conversely, broadcasting a rollout advantage to every generated token can introduce credit-assignment noise for an unfinished suffix. These choices should be documented separately: the reward definition determines whether correctness information exists, whereas token masking and policy-loss aggregation determine how it reaches the update. The analytical results isolate the first issue and state explicit approximations for the second.

\clearpage
\section{Additional Reward and Budget Derivations}
\label{app:additional-theory}

\subsection{Heterogeneous answer probabilities}

The reward-noise calculation can retain the heterogeneity across problems. For a candidate with latent correctness probabilities $p_1,\ldots,p_N$, let $\bar p=N^{-1}\sum_i p_i$ and $v_p=N^{-1}\sum_i(p_i-\bar p)^2$. Under conditional independence,
\begin{align}
\Var(r_{\mathrm{acc}}\mid p)
&=\frac{1}{N^2}\sum_i p_i(1-p_i)\nonumber\\
&=\frac{\bar p(1-\bar p)-v_p}{N}.
\end{align}
The identity follows by expanding $v_p=N^{-1}\sum_i p_i^2-\bar p^2$. It explains why the $1/(4N)$ bound is loose when some problems are almost certainly correct or incorrect. It also separates Bernoulli observation noise from variation in mean trajectory quality across candidates. The latter provides the signal that group-relative centering must retain.

\subsection{Explicit length penalties versus a shared cap}

Consider a local edit that adds $\Delta\ell_i>0$ tokens and increases problem $i$'s success probability by $\Delta p_i$, holding the completed set fixed. With an explicit reward $r_{\mathrm{acc}}-\lambda L$, the expected change is
\begin{equation}
\Delta\E[R_\lambda]=\frac{\Delta p_i}{N}-\lambda\Delta\ell_i.
\end{equation}
Useful reasoning is discouraged whenever its accuracy gain is smaller than this fixed token tax. Under CCR, the same edit changes expected accuracy by $\Delta p_i/N$ if it displaces no completed answer. If it displaces a set $D$, the change becomes $(\Delta p_i-\sum_{k\in D}p_k)/N$. This comparison assumes that the local edit has no other effects on the remaining solutions; the shared autoregressive context can violate that simplification.

The two objectives assign different costs to extra reasoning. A shared cap assigns an opportunity cost through displaced correctness, while an explicit penalty assigns a prescribed cost to every additional token.

\subsection{Feasible completion and conditional signal}

Let $b_i$ be the tokens required to finish problem $i$ along a particular successful trajectory. Completing the whole group requires $\sum_i b_i\leq B$. Partial credit relaxes the need to complete the entire group before receiving any accuracy reward, but does not remove the need to finish at least one answer. If all candidates fail that condition, $p_0=1$ and Eq.~\eqref{eq:informative} yields zero probability of an informative accuracy group. If the candidate count distribution has at least two values with positive mass, independently sampled groups can have nonzero accuracy advantages even when no sample completes all $N$ tasks.

This makes ordering relevant: an unusually long first solution may block every later answer. Heterogeneous difficulty, cross-problem interference, and ordering rules should be measured along with completion rate. A fixed $B/N$ is only a rough allocation heuristic because CCR does not impose equal per-problem sub-budgets.

\clearpage
\section{Qualitative Trajectory Summaries}
\label{app:cases}

We present base and $N=3$ trajectory summaries for Qwen3-4B-Thinking-2507. The examples below are ordinary single-problem evaluations. Their lengths may exceed the training cap because the cap applies to RL rollouts, not to these evaluation traces. The descriptions are summaries, not verbatim complete model outputs. Cases were selected for illustration, so their correctness rates and length reductions should not be averaged into a new benchmark result.

\subsection{Torus--Sphere Tangency (AIME 2024)}
\label{case:qualitative_aime24_torus}

\paragraph{Problem.} A torus with major radius $6$ and minor radius $3$ is tangent to a sphere of radius $11$ along circles of radii $r_i$ and $r_o$. If $r_i-r_o=\frac{m}{n}$, find $m+n$. \quad Ground Truth: 127

\paragraph{Base checkpoint (32,312 tokens).}
Searches over possible sphere-center offsets $d$ and repeatedly checks special cases. Several candidate radii are rejected as invalid or irrational, but the trace never derives the correct tangency-circle formula. The reasoning remains dominated by coordinate trial-and-error and finally reports an incorrect value.

$\text{Answer: }\boxed{352}\ \times$

\paragraph{CCR, $N=3$ (17,837 tokens).}
Keeps the essential tangency geometry while compressing the derivation and avoids the long offset search. Uses the inner and outer tangency-circle formulas directly, then computes the difference cleanly.

\[
r_i-r_o
=
\frac{33}{4}-\frac{33}{7}
=
\frac{99}{28}.
\]

$\text{Answer: }\boxed{127}\ \checkmark$

\subsection{Date Digit Parity (AMC 2023)}
\label{case:qualitative_amc23_date_parity}

\paragraph{Problem.} A date in 2023 is displayed as an 8-digit integer YYYYMMDD. For how many dates in 2023 does every digit appear an even number of times? \quad Ground Truth: 9

\paragraph{Base checkpoint (25,269 tokens).}
Sets up the correct parity condition from the fixed year 2023: digits 0 and 3 must appear oddly in the month-day part, while all other digits must appear evenly. Performs a long exhaustive enumeration over valid dates and recovers all nine satisfying dates. The reasoning is correct but token-heavy.

$\text{Answer: }\boxed{9}\ \checkmark$

\paragraph{CCR, $N=3$ (15,316 tokens).}
Uses month-by-month parity vectors rather than broad enumeration. Finds the valid dates by month: Jan has 2, Feb has 1, Mar has 2, Oct has 2, and Nov has 2.

\[
2+1+2+2+2=9.
\]

$\text{Answer: }\boxed{9}\ \checkmark$

\subsection{Party Graph with Enemy Closure}
\label{case:qualitative_brumo25_party_graph}

\paragraph{Problem.} Bruno invites $n$ guests. Each pair of guests are either friends or enemies. Each guest has exactly $12$ enemies, and all guests believe that the friend of an enemy is an enemy. Find the sum of all possible values of $n$. \quad Ground Truth: 100

\paragraph{Base checkpoint (26,146 tokens).}
Works in the enemy graph and correctly finds two extremal examples: everyone mutually enemies, giving $n=13$, and two equal friend groups, giving $n=24$. However, it treats these examples as exhaustive after only informal testing. It misses the full disjoint-clique structure of the friend graph, so it loses the intermediate values $14,15,16,18$.

$\text{Answer: }\boxed{37}\ \times$

\paragraph{CCR, $N=3$ (14,139 tokens).}
Uses the global clique structure in a concise derivation. Lists the divisor cases directly:
\[
k-1\in\{1,2,3,4,6,12\}.
\]

Hence
\[
n\in\{13,14,15,16,18,24\}
\]
\[
13+14+15+16+18+24=100.
\]

$\text{Answer: }\boxed{100}\ \checkmark$

\subsection{Vector Projection (MATH-500)}
\label{case:qualitative_math500_projection}

\paragraph{Problem.} Find the projection of $\mathbf{a}$ onto $\mathbf{b}=\begin{pmatrix}2\\6\\3\end{pmatrix}$ if $\mathbf{a}\cdot\mathbf{b}=8$. \quad Ground Truth: $\begin{pmatrix}16/49\\48/49\\24/49\end{pmatrix}$

\paragraph{Base checkpoint (3,717 tokens).}
Treats ``projection'' as the scalar projection rather than the vector projection. Computes $\|\mathbf{b}\|=7$ and applies $\frac{\mathbf{a}\cdot\mathbf{b}}{\|\mathbf{b}\|}=\frac{8}{7}$. The reasoning notices the ambiguity but resolves it in the wrong direction.

$\text{Answer: }\boxed{\dfrac{8}{7}}\ \times$

\paragraph{CCR, $N=3$ (1,546 tokens).}
Also identifies that the requested projection should be the vector projection. Keeps the derivation compact:
\[
\frac{8}{49}
\begin{pmatrix}2\\6\\3\end{pmatrix}
=
\begin{pmatrix}
16/49\\48/49\\24/49
\end{pmatrix}.
\]

Despite briefly considering the scalar interpretation, it selects the correct vector-valued answer.

$\text{Answer: }\boxed{\begin{pmatrix}16/49\\48/49\\24/49\end{pmatrix}}\ \checkmark$

\subsection{Polytropic Atmosphere Hydrostatic Equilibrium (Minerva)}
\label{case:qualitative_minerva_polytropic}

\paragraph{Problem.} A thin stellar atmosphere obeys $P=K\rho^{5/3}$ with constant gravitational acceleration $g$. Using hydrostatic equilibrium, derive $P(z)$ above the surface, with $P(0)=P_0$. \quad Ground Truth: $\left[P_{0}^{2/5}-\frac{2}{5}gK^{-3/5}z\right]^{5/2}$

\paragraph{Base checkpoint (5,683 tokens).}
Starts from the correct hydrostatic equation $dP/dz=-\rho g$ and the correct substitution $\rho=(P/K)^{3/5}$. However, it mishandles the factor of $K$ during separation, effectively moving $K^{3/5}$ into the numerator. This yields the wrong pressure profile
\[
P(z)=\left[P_0^{2/5}-\frac{2gK^{3/5}}{5}z\right]^{5/2}.
\]

$\boxed{\left[P_0^{2/5}-\frac{2gK^{3/5}}{5}z\right]^{5/2}}\ \times$

\paragraph{CCR, $N=3$ (2,930 tokens).}
Keeps the essential hydrostatic derivation while compressing the solution. After substituting $\rho=(P/K)^{3/5}$, it explicitly writes the correct differential equation with $K^{-3/5}$.

\[
P^{2/5}
=
P_0^{2/5}
-
\frac{2g}{5K^{3/5}}z.
\]

A dimensional check reinforces the correct dependence on $K^{3/5}$ in the denominator.

$\boxed{\left[P_0^{2/5}-\frac{2g}{5K^{3/5}}z\right]^{5/2}}\ \checkmark$

\subsection{Three Non-overlapping Squares}
\label{case:qualitative_olympiad_squares}

\paragraph{Problem.} Three non-overlapping squares of positive integer side lengths each have one vertex at the origin and sides parallel to the coordinate axes. Together, the three squares enclose a region whose area is $41$. Compute the largest possible perimeter of the region. \quad Ground Truth: 32

\paragraph{Base checkpoint (24,806 tokens).}
Finds the two possible side-length triples $(1,2,6)$ and $(3,4,4)$, but repeatedly changes the perimeter formula. Even after computing $32$ for both triples, it overrides that result using an unsupported ``more precise'' boundary correction. The final answer is therefore driven by a late inconsistent shortcut rather than the geometric boundary calculation.

$\boxed{30}\ \times$

\paragraph{CCR, $N=3$ (17,823 tokens).}
Keeps the same finite case analysis but compresses the search. Correctly minimizes the total shared boundary: two shared edges of the smallest square.

\[
(1,2,6):\ 4\cdot 9-4=32,
\]
\[
(3,4,4):\ 4\cdot 11-12=32.
\]

$\boxed{32}\ \checkmark$

\subsection{Sphere Tangent to Six Tetrahedra}
\label{case:qualitative_olympiad_tetrahedra}

\paragraph{Problem.} Six solid regular tetrahedra are placed on a flat surface so that their bases form a regular hexagon $\mathcal{H}$ with side length $1$, and so that the top vertices are coplanar. A spherical ball of radius $r$ has center directly above the center of the hexagon, rests on the tetrahedra, and is tangent to one edge from each tetrahedron. If the ball's center is coplanar with the top vertices, compute $r$. \quad Ground Truth: $\frac{\sqrt{2}}{3}$

\paragraph{Base checkpoint (31,054 tokens).}
Sets up reasonable coordinates for the hexagon and the top-vertex height $h=\sqrt{2/3}$. Then selects a lateral edge from an outer base vertex to the top vertex, rather than the symmetric edge adjacent to the hexagon center. The point-to-line distance therefore becomes $\sqrt{11}/6$, a plausible but wrong radius.

$\boxed{\frac{\sqrt{11}}{6}}\ \times$

\paragraph{CCR, $N=3$ (17,030 tokens).}
Makes the closest-edge choice explicit: the tangent edge is $OT_k$, from the hexagon center to a top vertex. Uses a direct cross-product distance
\[
r=\frac{\lVert \vec{CO}\times \vec{T_kO}\rVert}{\lVert \vec{T_kO}\rVert}.
\]

The computation yields $\sqrt{2}/3$ without the baseline's lateral-edge detour.

$\boxed{\frac{\sqrt{2}}{3}}\ \checkmark$

\subsection{Trigonometric Cubic}
\label{case:qualitative_olympiad_trig_cubic}

\paragraph{Problem.} For real numbers $\alpha,B,C$, the zeros of $T(x)=x^3+x^2+Bx+C$ are $\sin^2\alpha$, $\cos^2\alpha$, and $-\csc^2\alpha$. Compute $T(5)$. \quad Ground Truth: $\frac{567}{4}$

\paragraph{Base checkpoint (32,631 tokens).}
Introduces an auxiliary variable and derives a long expression for $T(5)$, but never proves it is constant. Switches to numerical sampling around $s\approx 1.9$ and observes values near $129$. The final answer is a rounded numerical guess rather than a consequence of Vieta's formulas.

$\boxed{129}\ \times$

\paragraph{CCR, $N=3$ (21,204 tokens).}
Keeps the product form
\[
T(5)=(5-r_1)(5-r_2)(5-r_3)
\]
and substitutes the root triple $(1/2,1/2,-2)$. The arithmetic is direct and yields $\frac{567}{4}$.

$\boxed{\frac{567}{4}}\ \checkmark$

\subsection{Product of Factorials}
\label{case:qualitative_olympiad_factorials}

\paragraph{Problem.} Given positive integers $a,b,c,d$ such that $a!\cdot b!\cdot c!=d!$ and $a+b+c+d=37$, compute $abcd$. \quad Ground Truth: 2240

\paragraph{Base checkpoint (32,643 tokens).}
Performs a long unstructured enumeration of many quadruples and known factorial identities. Finds several near misses but fails to reorganize the search around the ratio $d!/c!$. The trace reaches the token limit without a final answer.

$\text{No final answer}\ \times$

\paragraph{CCR, $N=3$ (15,458 tokens).}
Checks $d=16$ early and computes
\[
\frac{16!}{14!}=240=2!\cdot 5!.
\]

Verifies both constraints
\[
2!5!14!=16!,\qquad 2+5+14+16=37.
\]

Then computes $abcd=2\cdot 5\cdot 14\cdot 16=2240$.

$\boxed{2240}\ \checkmark$